\documentclass[conference]{IEEEtran}

\usepackage{cite}
\usepackage{graphicx}
\usepackage{array}
\usepackage{booktabs}
\usepackage{textcomp}
\usepackage{url}
\usepackage{microtype}
\usepackage{amstext}
\usepackage[scientific-notation=true]{siunitx}
\def\BibTeX{{\rm B\kern-.05em{\sc i\kern-.025em b}\kern-.08em
    T\kern-.1667em\lower.7ex\hbox{E}\kern-.125emX}}
    
\def\mybar#1{
  \color{black}
  \rule{#1cm}{6pt}
}

\makeatletter

\def\ps@IEEEtitlepagestyle{
  \def\@oddfoot{\mycopyrightnotice}
  \def\@evenfoot{}
}
\def\mycopyrightnotice{
  {\footnotesize $\copyright$ 2018 IEEE \hfill}
  \gdef\mycopyrightnotice{}
}
    
\begin{document}
\bstctlcite{IEEEexample:BSTcontrol}

\title{Multi-label Classification of Surgical Tools with \\ Convolutional Neural Networks}

\author{
\IEEEauthorblockN{Jonas Prellberg}
\IEEEauthorblockA{
\textit{Dept. of Computer Science}\\
\textit{University of Oldenburg}\\
jonas.prellberg@uni-oldenburg.de}
\and
\IEEEauthorblockN{Oliver Kramer}
\IEEEauthorblockA{\textit{Dept. of Computer Science} \\
\textit{University of Oldenburg}\\
oliver.kramer@uni-oldenburg.de}
}

\maketitle

\begin{abstract}
Automatic tool detection from surgical imagery has a multitude of
useful applications, such as real-time computer assistance for the
surgeon. Using the successful residual network architecture, a system that can
distinguish 21 different tools in cataract surgery videos is created.
The videos are provided as part of the 2017 CATARACTS challenge and
pose difficulties found in many real-world datasets, for example a
strong class imbalance. The construction of the detection system is guided by a
wide array of experiments that explore different design decisions.
\end{abstract}


\section{Introduction}

Analyzing medical image data is an active area of research
\cite{Zhou_2017_CVPR, Wang_2017_CVPR} that commonly deals with radiology
and histology images. However, surgeries in modern operating rooms
are assisted by imaging technologies as well. For example, video recordings of surgeries allow
to generate operation reports and reconstruct the surgical workflow \cite{cts-table-video}
or enable computer-assisted intervention systems that provide real-time recommendations
and warnings to the surgeon \cite{cts-stat-models}. The latter are
especially useful for inexperienced surgeons or surgeons in training.
Such systems should be vision-based and not rely on markers so that
ordinary surgical tools can be used and the system works in any operating
room \cite{tooldetection}.

Surgical tool detection in images can be used as the foundation for
higher-level systems, e.g. as in \cite{cts-stat-models} where surgical
steps and phases are inferred from tool presence. To stimulate research
in tool detection, the CATARACTS challenge\footnote{\url{https://cataracts.grand-challenge.org}}
has been organized by researchers
at the Laboratory of Medical Information Processing
in Brest. They provide a collection of annotated cataract surgery
videos with the goal to automatically detect the presence of 21 different
surgical tools in each video frame. A cataract is the clouding of
the eye's natural lens and leads to poor vision. During the surgery
the natural lens is destroyed and replaced with an artificial one.
In machine learning terms, the challenge is a supervised classification
problem on medical images.

Most previous work on surgical tool detection is performed with traditional
machine learning techniques and hand-crafted features \cite{tooldetection}.
In contrast, this paper explores multiple approaches based on convolutional
neural networks (CNN) to detect surgical tools in videos. CNNs have
significantly improved the state of the art in vision based problems
over the last years and therefore are a promising choice for this
challenge. All models presented in this paper are based on the successful
residual network (ResNet) \cite{resnet} because it is among the best
performing models in several vision challenges and implementations
pretrained on ImageNet \cite{ILSVRC15} are readily available in machine
learning frameworks. This paper focuses on CNNs that classify single
frames. While multi-frame models might improve the detection accuracy,
they are vastly more difficult to train \cite{Karpathy:2014:LVC:2679600.2680211},
which is problematic since some classes in the CATARACTS dataset have very few training examples.

In this paper a wide array of design choices is explored to create
an effective CNN for surgical tool detection.
First, different ways to use ResNet for transfer learning are compared.
The best results are achieved by fine-tuning ResNet while keeping
some layers fixed. On the other hand, using ResNet as a feature extractor
performs worse than fine-tuning in all experiments.
Second, the traditional CNN design for supervised learning is compared
to a design proposed by Oquab et al. \cite{weak-sup-location} for weakly supervised
learning. As will be explained, images from the CATARACTS dataset can
contain multiple objects, but bounding box information is not available. Therefore,
the problem can also be cast as a weakly-supervised learning problem. However, experiments
show that networks following the traditional design perform better
in all cases.
Third, a weighted loss function is explored to combat the class imbalance
found in the dataset. This leads to slight generalization improvements.
The final network achieves 0.957
AUC\footnote{AUC is the area under the receiver operating characteristic (ROC)
curve.} on the challenge's test set. 

The remainder of this paper is organized as follows: In Section~\ref{sec:related-work}
related literature that was helpful for tackling this challenge is
reviewed. Section~\ref{sec:network-architectures} presents two families
of network architectures to detect surgical tools in video frames.
These networks will be used for experiments and compared against each
other. Afterwards, in Section~\ref{sec:dataset}, the CATARACTS dataset
is described. The difficulties inherent to this dataset and the approaches
to solve them are explained. Finally, in Section~\ref{sec:experiments},
a wide array of experiments is conducted to explore possible design choices, finally resulting
in the classifier model. Section~\ref{sec:conclusion}
summarizes the paper.

\section{Related work\label{sec:related-work}}

\textbf{Convolutional neural networks.} CNNs dominate recent computer vision challenges and
especially residual networks \cite{resnet} are among the most successful architectures.
Residual networks were the first deep architectures that did
not suffer from a performance degradation due to the increased difficulty
of optimizing a large number of layers. This is achieved by adding
regularly spaced shortcut connections to an otherwise linear stack
of layers.

\textbf{Object detection.} Object detection is the problem of detecting
possibly multiple objects of various classes in an image and returning
their locations. This is somewhat related to the problem at hand:
surgical tools must be detected in an image but the location is irrelevant.
However, all state-of-the-art object detectors reviewed in \cite{DBLP:journals/corr/HuangRSZKFFWSG016} need bounding box labels for their training,
which are not available in the CATARACTS dataset. Consequently, existing
object detection networks are unsuitable for this problem.

\textbf{Multi-label image classification.} In the dataset at hand zero to three
surgical tools can be visible in each video frame. Therefore, one
image is associated with a set of labels instead of exactly one label.
This setting is called multi-label learning. Using neural networks for multi-label problems
is mostly a matter of choosing an appropriate last layer, loss function
and (often implicitly) problem transformation. Specifically for CNNs,
Gong et al. \cite{DBLP:journals/corr/GongJLTI13} compare 
different loss functions in a multi-label setting. Wei et al. \cite{weak-sup-hcp}
and Oquab et al. \cite{weak-sup-location} both perform multi-label image classification by treating the task as $c$
independent binary classification problems. As such, their CNNs have
$c$ output nodes which is an implicit way of applying the binary
relevance transformation \cite{multi-label}.

\textbf{Transfer learning.} Training CNNs with millions of weights requires a large amount of labeled training data. This is expensive or even infeasible to collect for many tasks. However, it has been shown that the lower layers of deep
CNNs used for image classification learn very general filters that
are applicable to different datasets as well \cite{Yosinski:2014:TFD:2969033.2969197}.
The idea behind transfer learning is to pretrain a network on a very large image datasets, such as ImageNet. The network can
then be used as a fixed feature extractor or fine-tuned by
training it further on the target dataset. Both of these transfer
learning approaches will be explored and compared in this paper.

\textbf{Imbalanced datasets.} The CATARACTS dataset is drastically imbalanced
since some tools are used much more often during cataract
surgery than others. This makes classification accuracy meaningless
as a performance metric. Receiver operating characteristic (ROC)
graphs \cite{roc-introduction} are a viable alternative that can be aggregated
into a scalar performance metric by calculating the area under curve (AUC).

Furthermore, the class imbalance influences a neural network's stochastic
gradient descent training. Since training examples of the majority
classes are much more common, the gradient direction is dominated
by examples of these majority classes \cite{imbalanced-data-review,Mazurowski2008}.
A possible remedy is to scale the loss by class prevalence, i.e. increasing
the loss and therefore gradient length for low-prevalence classes.
This approach is explored in this paper as well.

\textbf{Surgical tool detection. }Much of the work related to visual
surgical tool detection from 2000 to 2015 has been reviewed in \cite{tooldetection}.
While the goals vary from detecting presence to estimating tool poses
in 3D, all 28 reviewed methods have in common that they use hand-crafted
image features as input to different modeling strategies.
The same holds true for more recent publications
\cite{cts-cbvr,cts-polynomials,cts-table-video,cts-stat-models} that
also work with cataract surgery datasets. Additionally, \cite{cts-stat-models}
employs preprocessing methods specifically tuned to the cataract surgery dataset that
prevents easy application to different tool detection problems. Work using CNNs in this 
application domain is just beginning to emerge. EndoNet \cite{endonet} is similar to this
work: AlexNet is fine-tuned to detect different tools and surgical phases in laparoscopic
videos. The main difference is a three-times smaller amount of tools in their dataset.
Furthermore, there is concurrent work with CNNs being done by other authors for the CATARACTS challenge.

\section{Network architectures\label{sec:network-architectures}}

\begin{figure}
\begin{centering}
\includegraphics[width=0.95\columnwidth]{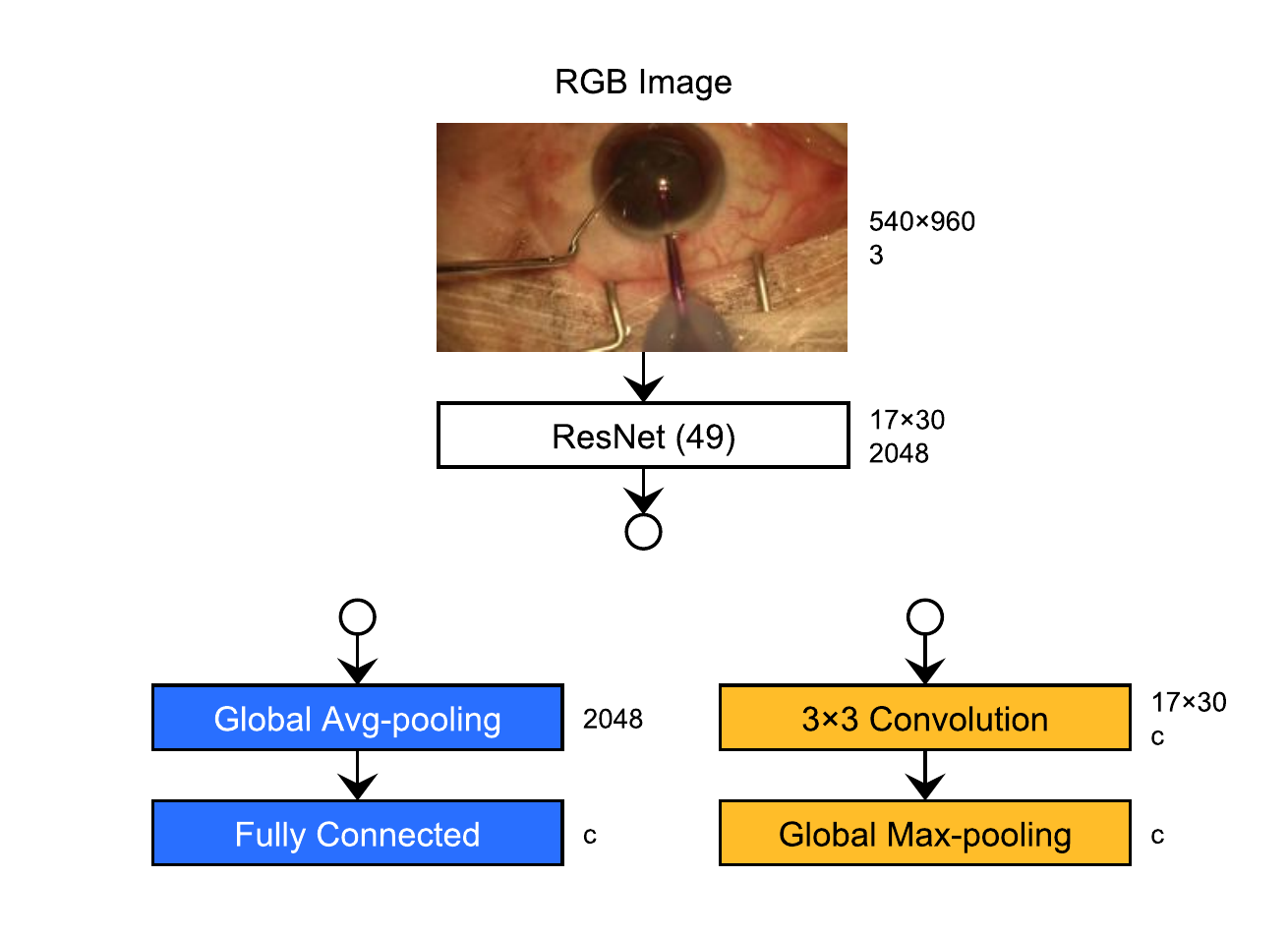}
\par\end{centering}
\caption[Fine-tuning (FT) network family]{Fine-tuning (FT) network family. The figure shows two possible instances.
The output of the 49th ResNet layer is either fed into the \emph{avg-fc}
classification head (left) or the \emph{conv-max} classification head
(right). The labels next to each node are feature map size and feature
map count. The parameter $c$ denotes the number of classes.\label{fig:network-ft}}
\end{figure}

All network architectures presented in this section build upon the
50-layer ResNet \cite{resnet} because of its successes in various
computer vision challenges. While larger variants of ResNet with more layers
might yield improved performance, their training cost was prohibitive given the time
and hardware constraints for this work. Two families of networks will be described
from which several instances of networks can be derived. The first
family is designed for fine-tuning experiments, while the second one
uses ResNet as a fixed feature extractor. All instances are linear
stacks of multiple network parts: first some layers of ResNet, then
optionally some custom layers and finally the classification head.
There are two choices for the classification head: \emph{avg-fc} and
\emph{conv-max} (see Figure~\ref{fig:network-ft}).

The \emph{avg-fc} case is the standard supervised classification setup
of ResNet. The feature maps created by the last convolutional layer
are global average-pooled and then fed into a fully connected layer
with as many units as there are classes. This approach is also found
in other popular network architectures such as Inception \cite{inception}
and is therefore used as the baseline in all experiments.

The \emph{conv-max} case is the setup proposed by Oquab et al. \cite{weak-sup-location}
for weakly supervised multi-label learning. In their work AlexNet is used as a
fixed feature extractor, followed by two trainable convolutional ``adaption''
layers and finally a global max-pooling. Because the network family
here uses fine-tuning instead of a feature extractor with fixed weights,
adaption layers in their original sense are unnecessary. However,
one convolutional layer is still employed so that there are as many
feature maps as classes before applying the global max-pooling.
The paper claims that the global max-pooling ``effectively searches
for the best-scoring candidate object position within the image, which
is crucial for weakly supervised learning where the exact position
of the object within the image is not given at training'' \cite{weak-sup-location}.
Since the CATARACTS dataset also poses a multi-label problem and has no bounding-box
information, this approach seems well suited to the problem and will
be compared to the traditional setup.

Both variants have $c$ outputs with $c$ being the number of unique
surgical tools to detect. Also, the activation function of the last
fully connected or convolutional layer respectively is the sigmoid
function to squash values into the range $\left[0,1\right]$. Note
that this is different from a softmax function because the $c$ outputs are independent
and generally do not sum to one.

All networks presented here use transfer learning, either in the form of fine-tuning or with a fixed feature extractor.
Transfer learning is essential for good performance on the CATARACTS dataset because some
classes contain very few training examples. This claim will be supported empirically in Section~\ref{subsec:validation-set-results}.

The network family for fine-tuning (FT) experiments is illustrated
in Figure~\ref{fig:network-ft}. It uses all 49 convolutional layers
of ResNet. The output feature maps of the 49th convolutional layer
are either fed into the \emph{avg-fc} or \emph{conv-max} classification
head. In general, all weights in this network family are trainable
but for some experiments the first $k$ layers of ResNet will be frozen.
The pretrained weights provide a form regularization that will become
stronger with an increasing amount of frozen layers. In order to optimize
for generalization performance, different settings are explored.

Training this architecture requires a significant amount of memory
due to the many trainable weights and input image resolution. All
explored instances of this fine-tuning network family vary only in
the amount of ResNet layers that are frozen and the type of classification
head.

\begin{figure}
\begin{centering}
\includegraphics[width=0.78\columnwidth]{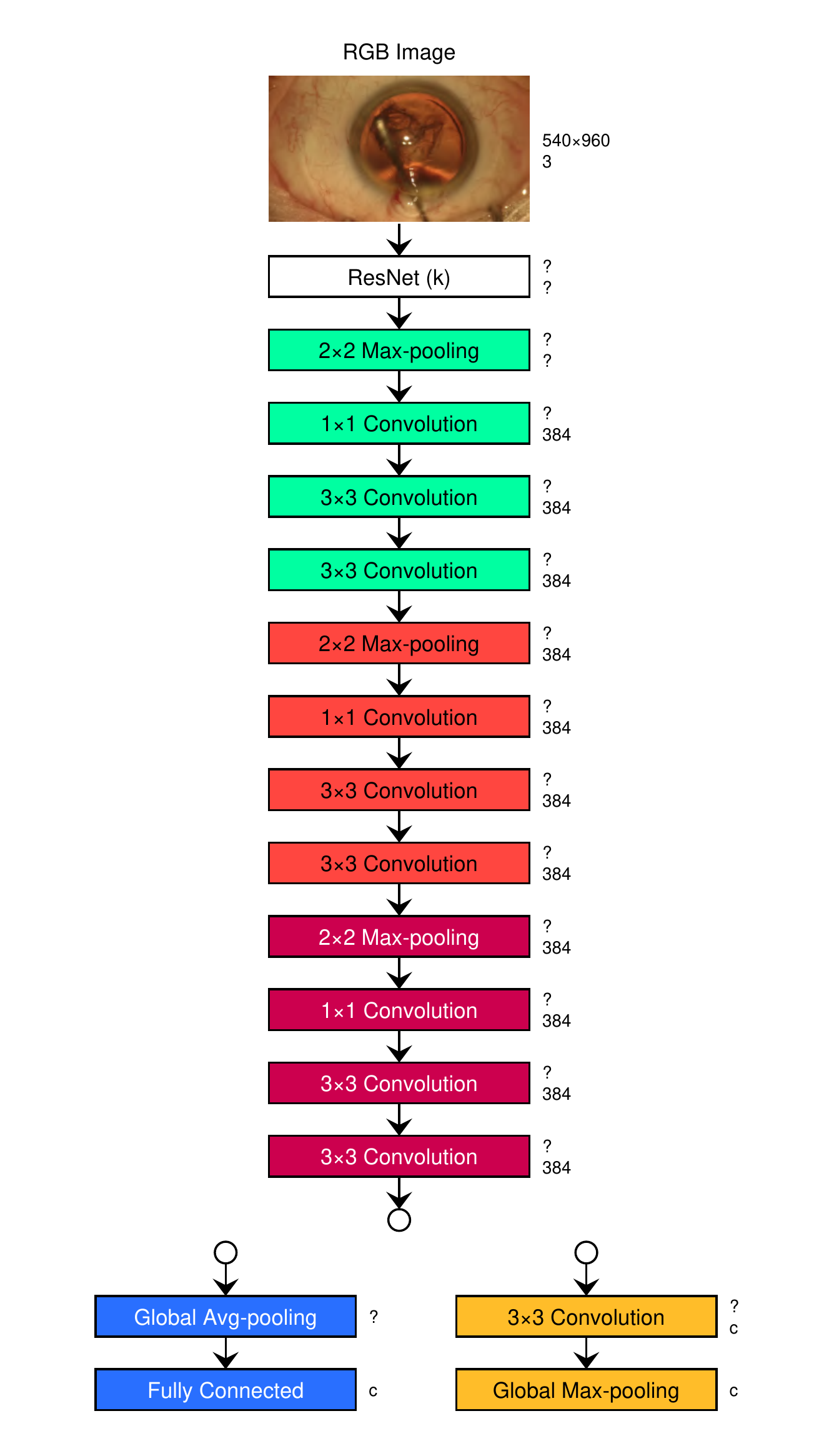}
\par\end{centering}
\centering{}\caption[Fixed feature extractor (FFE) network family]{Fixed feature extractor (FFE) network family. The figure shows 8
different possible instances. The input image is fed into a fixed
weight ResNet with $k\in\left\{ 22,31,40,49\right\} $ layers. The
resulting feature maps are processed by a custom architecture. Finally,
the result is either fed into the \emph{avg-fc} classification head
(bottom left) or the \emph{conv-max} classification head (bottom right).\label{fig:network-ffe}}
\end{figure}

\begin{figure}
\begin{centering}
\includegraphics[width=0.95\columnwidth]{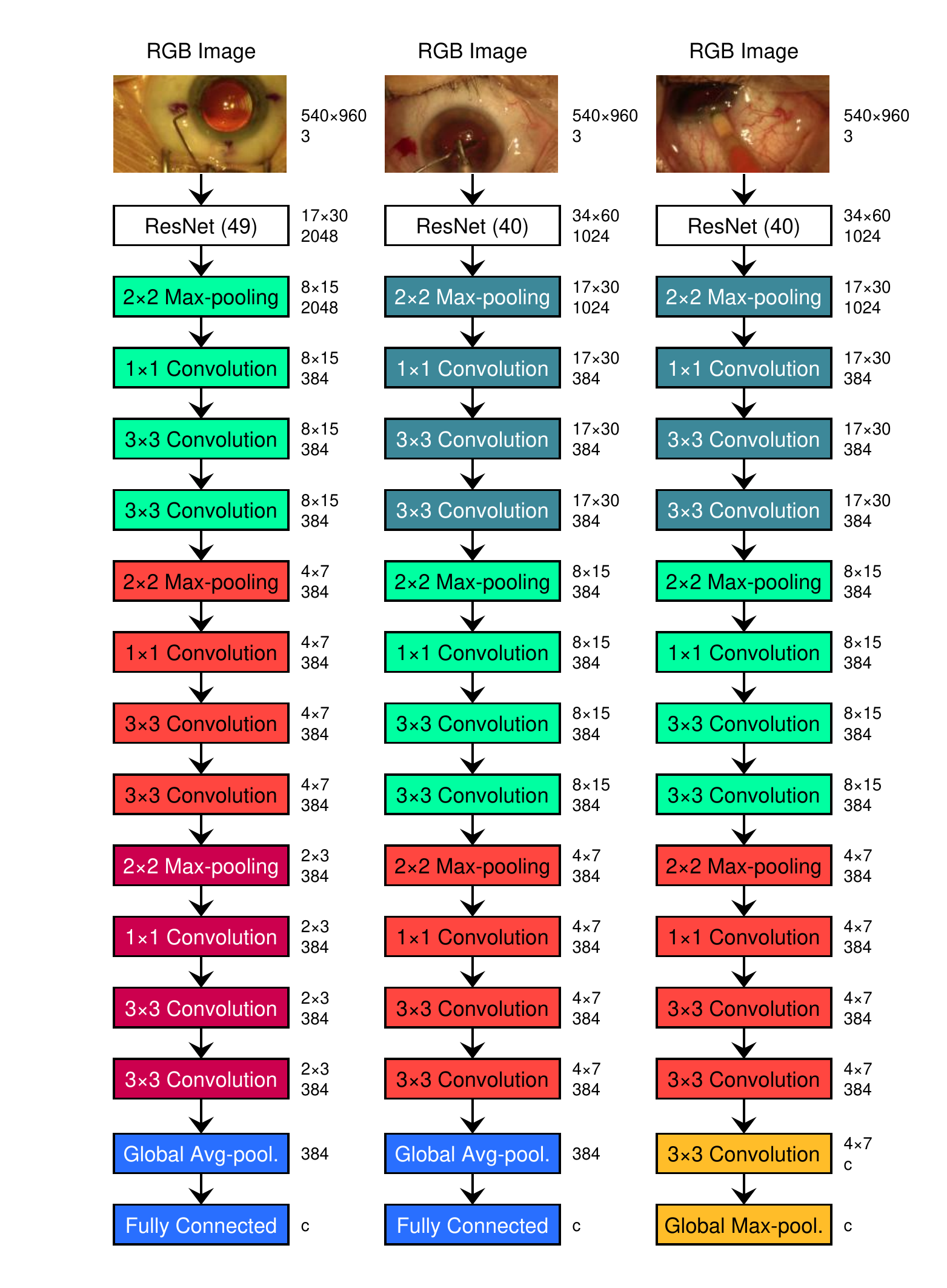}
\par\end{centering}
\caption[Three exemplary instances of the FFE network family]{Three exemplary instances of the FFE network family with different
choices for $k$ and the classification head. Left: $k=49$, \emph{avg-fc}.
Center: $k=40$, \emph{avg-fc}. Right: $k=40$, \emph{conv-max}.\label{fig:network-ffe-examples}}
\end{figure}

The network family for fixed feature extractor (FFE) experiments is
illustrated in Figure~\ref{fig:network-ffe}. The first $k$ layers
of ResNet are used as a feature extractor with fixed weights. The
resulting feature maps are fed to a max-pooling layer and three layers
of convolutions with 384 feature maps each. Each convolution is followed
by Batch Normalization \cite{batchnorm} and a ReLU activation function.
The same structure is repeated two more times. This specific custom
architecture was found to be the best in preliminary experiments but
is not necessarily the best possible one. Like for the first network
family, the final part is either the \emph{avg-fc} or \emph{conv-max}
classification head. Training this architecture is relatively inexpensive
in terms of required memory because all ResNet weights are fixed and
only the custom layers are trained. The explored instances of this
family vary in the amount of ResNet layers that are used as a FFE
and the type of classification head. For clarity, Figure~\ref{fig:network-ffe-examples}
shows three specific instances of the FFE family. Note how the feature
map resolution changes depending on the amount of ResNet layers used
as a feature extractor.

In both network families each output node corresponds to a predicted
score for one of the $c$ surgical instruments. Because the instruments
are not mutually exclusive, the binary relevance transformation is
used so that the task is treated as $c$ separate binary classification
problems. For each of these binary problems the training dataset can
be partitioned into a set of positive and negative examples. The set
of positive examples for class $i$ contains all frames that show
tool $i$ (potentially together with other tools), while the set of
negative examples contains all other frames.

Each of the binary problems is trained using the cross-entropy loss
function
\begin{equation}
H\left(p,q\right)=-\left(1-p\right)\log\left(1-q\right)-p\log\left(q\right)\,,
\end{equation}
where $p\in\left\{ 0,1\right\} $ is the ground-truth probability
for the positive class and $q\in\left[0,1\right]$ is the predicted
probability (network output) for the positive class. While the binary
problems are conceptually separate, this makes no difference during
training. For each example in the training set, the loss is calculated
for all $c$ network outputs and then backpropagated through the network.

\section{Dataset\label{sec:dataset}}

\subsection{Dataset description}

During 50 cataract surgeries the view of the surgeon's microscope
has been video recorded in a resolution of 1920$\times$1080 pixels
at 30 frames per second. The viewpoint is mostly static but the camera
shakes and occasionally the zoom level is changed. On average, the
surgery duration is 11 minutes so that over 9 hours of video are available.
A total of 21 different surgical tools are used in the videos and
up to 3 tools can be visible at a time. However, this is extremely
rare (0.04\,\%) and most frames show no tools (45\,\%), one tool
(38\,\%) or two tools (17\,\%). The videos have been independently
annotated by two experts, which allows to ignore frames where the
experts disagree during the evaluation. For each frame and surgical
tool, a label indicates if the tool touches the eyeball. Note that
this means a tool can be visible but still not annotated because it
does not yet touch the eyeball. Also, multiple tools can be present
in a single frame but bounding box information to distinguish the
tools in the frame is not available. This is essentially a weakly
supervised classification setting or more specifically multi-label
classification. Finally, all 50 videos have been evenly split into
a training and test set but labels are only provided for the training
set.

\subsection{Dataset challenges and preprocessing}

The dataset poses some challenges that have to be addressed before
being usable for training. First of all, the video resolution is very
high at 1920$\times$1080 pixels. This is a problem for CNNs as the
required processing time and especially memory is directly influenced
by the image resolution. However, the images cannot be scaled down
arbitrarily because most tools are elongated objects of only about
30 pixels width. A resolution of 960$\times$540 pixels was found
to be a good compromise between resource demands and object size.

Due to the nature of video, subsequent frames are extremely similar
to one another. They are heavily correlated and using neighboring
frames during supervised learning yields almost no information gain.
Therefore, only every sixth frame of each video is used during training
which leads to 200\,ms intervals between processed frames. No significant
information is lost this way, but the required training time is reduced
considerably.

Additionally, the similarity between neighboring frames has implications
on the choice of validation data. Consider a random split of all frames
into a large training and small validation set. It is highly likely
that for each frame in the validation set either the predecessor or
successor of that frame is part of the training set. The validation
error would therefore be a significant underestimation of the test
error. Instead, the training-validation-split is performed on the
video level: 5 of the 25 training videos are set aside for validation
purposes.

\begin{figure}
\noindent \begin{centering}
\includegraphics[width=1\columnwidth]{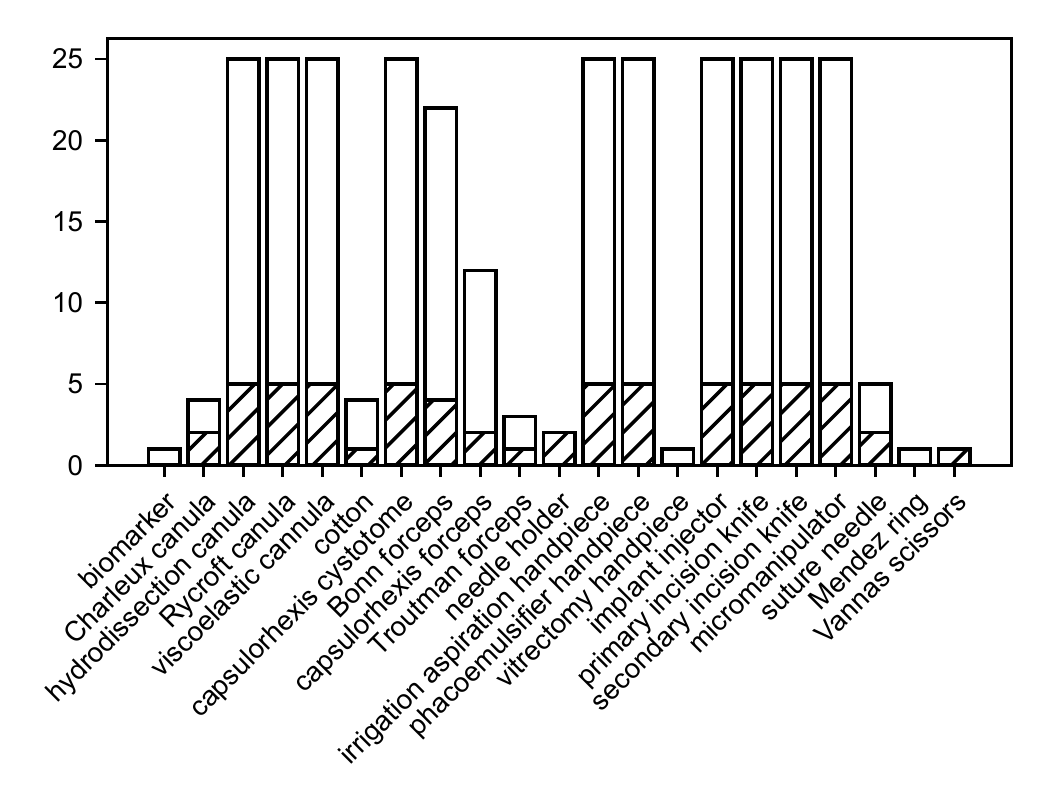}
\par\end{centering}
\caption[Training/validation set split]{Amount of videos showing each tool inside the training set (empty
bars) and validation set (striped bars) \label{fig:train-val-split}}
\end{figure}

The dataset also exhibits a strong class imbalance. Because almost
half of all frames do not show any tools, this subset of the data
is undersampled to 40\,\% of its original size. While the number
of available training examples for each tool also varies considerably,
the more important consideration is how many videos contain a sequence
showing each tool. This is relevant because frames showing the same
tool in different videos are a lot more varied than frames showing
the same tool in a single video. In consequence, the available videos
are distributed between training and validation set so that both contain
some videos showing each tool. The distribution of the eventual split
can be seen in Figure~\ref{fig:train-val-split}. Unfortunately,
it is only possible to split the videos in ways that have some tools
appearing only in the training \emph{or} validation set. This necessitates
to exclude the tools Vannas scissors, Mendez ring, vitrectomy handpiece,
needle holder and biomarker from the experiments for validation purposes.
Note that even for classes that appear in the same amount of videos,
there are still great differences in the amount of available training
examples. For example, the micromanipulator and implant injector are
both present in every video but the former is used multiple times
and for long periods per video while the latter is used exactly once
for a few seconds per video.

The resulting training set contains about 43k images and 16 different
classes. The validation set is comprised of all frames from the validation
videos with a total of 73k images. The validation set is only larger
than the training set because frames are skipped during training but
not during validation.

\subsection{Data augmentation}

Because some tools only appear in a single training video the variability
between the frames showing such tools is low. To tackle this problem,
several image augmentation techniques are employed. First, the image
is randomly cropped to 960$\times$540 by scaling the original image
down to 1024$\times$604 and sampling horizontal and vertical offsets
from a uniform distribution between 0 and 64. Afterwards, it is horizontally
flipped in 50\,\% of all cases. Then color augmentation as described
by \cite{alexnet} is applied with $\alpha$ drawn from a standard
normal distribution. Finally, the image is rotated around its center
by a value chosen from a random uniform distribution between -15\textdegree{}
and 15\textdegree .

\section{Experiments\label{sec:experiments}}

\subsection{Training and evaluation procedure\label{subsec:train-eval-procedure}}

As usual for image data, pixel values are scaled to be in $\left[0,1\right]$
and then zero-centered by subtracting the mean pixel activations over
the training set. In practice this involves calculating the average
value for each pixel and color channel over the training set. The
result is a ``mean image'' of the training set that can be subtracted
from each training example.

All networks are trained with stochastic gradient descent and with
0.9 momentum. The learning rate starts at 0.05 in all cases and decays
over the course of the training. The learning rate at each batch $n$
is computed as $0.05\,/\left(1+dn\right)$ with decay factor $d$.

When training a FFE network the decay factor is $d_\text{FFE} = \num{0.001}$,
the batch size is 32 and the training is run for 20k iterations. When
fine-tuning, the decay factor $d_\text{FT} = \num{0.000125}$ is chosen,
the batch size is 8 and the training is run for 60k iterations. The
amount of training iterations was chosen in such a way that the validation
AUC has already stagnated before finishing the training. In the fine-tuning
case the batch size is that much smaller than in the FFE case because
training all of ResNet's weights requires more memory than only training
the few custom layers after a FFE ResNet. It should be noted that
such a small batch size is not ideal when using batch normalization,
because the batch statistics vary much more. However,
the training of all networks still works well enough.

Experimental results are first obtained using the validation set as
presented in Figure~\ref{fig:train-val-split}. Therefore, only 16
classes are distinguished instead of 21. The assumption is that networks
that perform best on the validation set will also perform best on
the challenge's test set even though the networks will change slightly
to accommodate the additional class outputs. For the final submission
to the challenge, the best network is trained on all 25 training videos
without a validation set. Training without a validation set is necessary
because some tools are only present in a single video so that such
a tool is either part of the training set \emph{or} the validation
set. This ``blind'' training procedure is the main motivation to
use a static learning rate schedule because dynamic rules like reducing
the learning rate when the AUC plateaus are not possible without a
validation set.

The area under the ROC curve (AUC) is the performance metric used
in all experiments. Each of the network's $c$ outputs is a score
in $\left[0,1\right]$. A threshold is applied to decide whether that
score means a tool is present or not. By varying this threshold the
ROC curve is created. It is calculated separately for each of the
$c$ outputs because it is designed for binary classification problems.
Finally, the resulting values are averaged to get an aggregating performance
metric for the whole network. This is sometimes called macro-average
ROC.

\subsection{Results on validation set}\label{subsec:validation-set-results}

\textbf{Comparison of transfer learning settings.} The first series
of experiments is performed to compare different transfer learning
settings. Four instances of the FT and FFE family are trained using
the standard classification head \emph{avg-fc.} The FT instances vary
in the amount of ResNet layers that are frozen during training while
the FFE instances vary in the amount of ResNet layers that are used
as the feature extractor. The results are shown in Table~\ref{tab:results-ft}
and Table~\ref{tab:results-ffe} in column \emph{avg-fc}.

\begin{table}
\caption{Experimental results for FT networks\label{tab:results-ft}}
\centering{}%
\begin{tabular}{llll}
\toprule 
\multicolumn{2}{c}{\vspace{1mm}Model configuration} & \multicolumn{2}{c}{Val. AUC}\tabularnewline
{\scriptsize{}Name} & {\scriptsize{}Frozen layers} & {\scriptsize{}$\overline{\text{avg-fc}}$} & {\scriptsize{}conv-max}\tabularnewline
\midrule
\midrule 
FT0 & None & 0.9488 & 0.9225\tabularnewline
\addlinespace[1mm]
FT22 & Partial (22) & \textbf{0.9522} & 0.9135\tabularnewline
\addlinespace[1mm]
FT31 & Partial (31) & 0.9503 & 0.8909\tabularnewline
\addlinespace[1mm]
FT40 & Partial (40) & 0.9440 & 0.8742\tabularnewline
\bottomrule
\addlinespace[1mm]
\end{tabular}
\end{table}

\begin{table}
\caption{Experimental results for FFE networks\label{tab:results-ffe}}
\centering{}%
\begin{tabular}{lllll}
\toprule 
\multicolumn{3}{c}{\vspace{1mm}Model configuration} & \multicolumn{2}{c}{Val. AUC}\tabularnewline
{\scriptsize{}Name} & {\scriptsize{}ResNet $k$} & {\scriptsize{}Custom part} & {\scriptsize{}avg-fc} & {\scriptsize{}conv-max}\tabularnewline
\midrule
\midrule 
FFE22 & 22 & Yes & 0.9073 & 0.8304\tabularnewline
\addlinespace[1mm]
FFE31 & 31 & Yes & 0.9285 & 0.9187\tabularnewline
\addlinespace[1mm]
FFE40 & 40 & Yes & 0.9365 & 0.9266\tabularnewline
\addlinespace[1mm]
FFE49 & 49 & Yes & \textbf{0.9409} & 0.9180\tabularnewline
\addlinespace[1mm]
FFE49NC & 49 & No & 0.8161 & \textemdash{}\tabularnewline
\bottomrule
\addlinespace[1mm]
\end{tabular}
\end{table}

The FT instances perform consistently better than all FFE instances.
This is plausible as the target dataset is very different from ImageNet
so that the more flexible fine-tuning instances should adapt better.
Because the FT instances perform best, each of those four experiments
is repeated a total of five times with random initializations and
averaged to get more precise results. This allows to select the best
instance for further experiments and will also allow to quantify improvements
with greater certainty.

Among the FT instances a medium amount of frozen layers works best.
When freezing either 22 or 31 layers of ResNet the results improve
over freezing none or 40 layers. Both of these instances, FT22 and FT31,
perform similarly well at 0.95 AUC though FT31 is cheaper
to train because it has more frozen layers. Compared to FFE22 and FFE31,
which have as many total ResNet layers as the FT22 and FT31 networks have frozen layers,
fine-tuning performs better. There are two likely reasons for this:
First, the trainable custom part appended to the FFE
networks is of fixed, small capacity. In contrast, the trainable part of FT networks is the
complete remainder of ResNet that has not been frozen. Therefore, more
trainable capacity is available. Second, the weights in the FFE network's custom part are randomly
initialized, while the FT network's weights benefit from improved initial conditions due to pretraining.

Among the FFE instances, the performance rises with the amount of
ResNet layers that are used as the feature extractor. Interestingly,
the experiment FFE49 which uses the full ResNet as a feature extractor
achieves the best result
with 0.94 AUC despite the dataset differences. However, this is solely
due to the trainable custom layers inserted between the feature extractor
and the classification head. In the experiment denoted as FFE49NC
the trainable custom part is not included and the achieved performance
drops significantly to 0.82 AUC.

To test how important the pretraining on ImageNet is, the completely
trainable FT0 network is trained again starting from
a random weights initialization. The results are shown in Table~\ref{tab:results-pretraining}.
When training for the same number of iterations as FT0 a
significantly lower AUC of 0.92 is achieved. Even when training the
network until the validation AUC stagnates (100k iterations), the
final performance is still much lower than what is possible with pretraining.
In conclusion, the pretraining is vital for good performance.

It is also of interest if the same performance can be achieved with
lower resolution input images. Such a result would be very helpful
as the training times and memory requirements would lower dramatically.
The FT31 network is trained again with different configurations
on 480$\times$270 pixel input images but no AUC better than 0.93
could be achieved. Consequently, the resolution is not lowered.

\begin{table}
\caption{Influence of pretraining on ImageNet\label{tab:results-pretraining}}
\centering{}%
\begin{tabular}{llc}
\toprule 
\multicolumn{2}{c}{\vspace{1mm}FT0 model} & \multicolumn{1}{c}{Val. AUC}\tabularnewline
{\scriptsize{}Pretrained} & {\scriptsize{}Iterations} & {\scriptsize{}avg-fc}\tabularnewline
\midrule
\midrule 
Yes & 60k & \textbf{0.9488}\tabularnewline
\addlinespace[1mm]
No & 60k & 0.9180\tabularnewline
\addlinespace[1mm]
No & 100k & 0.9224\tabularnewline
\bottomrule
\addlinespace[1mm]
\end{tabular}
\end{table}

\textbf{Comparison of classification heads.} A second series of experiments
is conducted to compare between the two possible classification head
types: standard global average-pooling followed by a fully connected
layer (\emph{avg-fc}) and the proposal for weakly supervised learning
by Oquab et al. \cite{weak-sup-location} using a fully convolutional network and
a global max-pooling layer (\emph{conv-max}). The previous FT and
FFE experiments are repeated with the \emph{conv-max} classification
head and the results are shown in Table~\ref{tab:results-ft} and
Table~\ref{tab:results-ffe} in column \emph{conv-max}.

The \emph{conv-max} classification head performs consistently worse
than the standard \emph{avg-fc} approach in FT and FFE settings.
Oquab et al. \cite{weak-sup-location} report good performance on PASCAL VOC 2012
using their approach, which is specifically motivated by the multi-label, no bounding-box nature of their dataset. Since the CATARACTS dataset shares these characteristics, the approach should
be well suited but unexpectedly fails.
A possible explanation lies in the difference between the CATARACTS
and PASCAL VOC 2012 datasets. The average amount of classes per image
is 1.0 for CATARACTS (after applying the preprocessing steps) and
2.4 for PASCAL VOC 2012. Furthermore, the CATARACTS dataset contains
at least some frames for each tool that show no other tools. To test
if these circumstances are responsible for the bad performance, the
training on CATARACTS is repeated only with frames that show at least
two tools, bringing the classes per image up to 2.0. Unfortunately,
this reduces the dataset in size a lot so that good results cannot
be expected. The FT31 network is trained again on the modified
dataset\textemdash once with each classification head. The \emph{avg-fc}
variant achieves 0.60 AUC, while the \emph{conv-max} variant does
not even learn properly and gets stuck on 0.50 AUC. Since \emph{avg-fc}
still performs better the assumption cannot be validated.

\textbf{Weighted loss function.} Due to the heavy class imbalance
in the dataset, minority classes might not be trained as well as majority
classes, which dominate the gradient direction. To verify if better
results can be achieved when actively dealing with the class imbalance,
a weighted loss function is employed. To this end, each class is associated
with a weight
\begin{equation}
w_{i}=\sqrt{\frac{\max\left\{ f_{j}\,|\,1\leq j\leq c\right\} }{f_{i}}}\,,
\end{equation}
where $f_{i}$ is the frequency with that class $i$ appears in the
dataset. Ignoring the square root for a moment, this associates a
weight of $x$ to each class that appears $1/x$ as many times as
the majority class. Because the frequency differences in this dataset
are quite drastic, the square root was applied to prevent huge weights
and therefore gradient descent steps. Before the backwards pass, the
cross-entropy loss associated with each output (class) $i$ is multiplied
by the class weight $w_{i}$. This leads to all weight updates during
backpropagation being scaled by the same factor.

The best FT instances, FT22 and FT31, are trained
again five times with the described weighted loss and the averaged
results are shown in Table~\ref{tab:results-weighted}.

\begin{table}
\caption{Experimental results with a weighted loss function\label{tab:results-weighted}}
\centering{}%
\begin{tabular}{lllc}
\toprule 
\multicolumn{3}{c}{\vspace{1mm}Model configuration} & \multicolumn{1}{c}{Val. AUC}\tabularnewline
{\scriptsize{}Name} & {\scriptsize{}Weighted} & {\scriptsize{}Iterations} & {\scriptsize{}$\overline{\text{avg-fc}}$}\tabularnewline
\midrule
\midrule 
FT22 & No & 60k & 0.9522\tabularnewline
\addlinespace[1mm]
FT22{*} & Yes & 25k & 0.9552\tabularnewline
\addlinespace[1mm]
FT31 & No & 60k & 0.9503\tabularnewline
\addlinespace[1mm]
FT31{*} & Yes & 25k & \textbf{0.9581}\tabularnewline
\bottomrule
\addlinespace[1mm]
\end{tabular}
\end{table}

During experiments with the weighted loss, overfitting is immediately
observed. A plausible explanation is that all class weights $w_{i}$
are greater or equal to one, which leads to larger gradient descent
steps in general. Therefore, the training may progress faster and
overfitting is observed earlier. Training the same networks for 25k
iterations instead of 60k leads prevents the overfitting. In the end,
the weighted loss improves the AUC for both networks slightly and
FT31* achieves the highest overall performance of 0.96
AUC.

\textbf{Regularization.} Because the previous experiments overfit
quickly, regularization might help. The networks are trained again
with L2 regularization of 1e-4 and 1e-5 on all weights and biases.
However, the performance stays similar or degrades so no further experiments
in this direction are conducted.

\textbf{Final network.} For clarity, a complete description of the
best performing network, FT31*, follows. The network consists
of the first 49 layers of ResNet followed by the \emph{avg-fc} classification
head. The first 31 convolutional layers of ResNet are frozen so that
their weights are not modified during training. The gradient of each
output node is scaled by the associated class weight as described
previously. The network is pretrained on ImageNet and then fine-tuned
for 25k iterations on CATARACTS. Other settings such as learning rate
or batch size follow the description in Section~\ref{subsec:train-eval-procedure}.

\subsection{Results on test set}

The best network found in all experiments on the validation set, FT31*,
is used to classify the challenge's test set. The average AUC over
all classes is 0.957 and a detailed evaluation broken down by class
is shown in Table~\ref{tab:results-test}. The test set AUCs by tool
are somewhat correlated to each tool's training subset size. For the
three classes irrigation/aspiration handpiece, phacoemulsifier handpiece
and micromanipulator the network achieves more than 0.99 AUC and these
are also the classes that have the most training examples. In contrast,
the biomarker detection is hardly better than chance and it is one
of the classes that appear in only a single training video. Additionally,
the biomarker appears in the fewest video frames in total. The combination
of not having enough data and simultaneously being assigned the largest
class weight probably leads to severe overfitting. It should
still be noted that some classes such as the Mendez ring perform well
despite having few training examples. In this case the likely reason
is that the Mendez ring looks very different from all other tools
and is therefore easy to distinguish.

\begin{table}
\caption{Results of FT31* on the test set\label{tab:results-test}}
\centering{}%
\begin{tabular}{lll}
\toprule 
Tool & \multicolumn{2}{l}{Test set AUC}\tabularnewline
\midrule
\midrule 
Average & 0.9568 & \mybar{0.9136}\tabularnewline
\midrule
secondary incision knife & 0.9978 & \mybar{0.9956}\tabularnewline
\addlinespace[1mm]
capsulorhexis cystotome & 0.9976 & \mybar{0.9952}\tabularnewline
\addlinespace[1mm]
phacoemulsifier handpiece & 0.9971 & \mybar{0.9942}\tabularnewline
\addlinespace[1mm]
vitrectomy handpiece & 0.9932 & \mybar{0.9864}\tabularnewline
\addlinespace[1mm]
irrigation/aspiration handpiece & 0.9913 & \mybar{0.9826}\tabularnewline
\addlinespace[1mm]
micromanipulator & 0.9912 & \mybar{0.9824}\tabularnewline
\addlinespace[1mm]
Rycroft canula & 0.9908 & \mybar{0.9816}\tabularnewline
\addlinespace[1mm]
capsulorhexis forceps & 0.9888 & \mybar{0.9776}\tabularnewline
\addlinespace[1mm]
primary incision knife & 0.9848 & \mybar{0.9696}\tabularnewline
\addlinespace[1mm]
Mendez ring & 0.9814 & \mybar{0.9628}\tabularnewline
\addlinespace[1mm]
suture needle & 0.9796 & \mybar{0.9592}\tabularnewline
\addlinespace[1mm]
cotton & 0.9759 & \mybar{0.9518}\tabularnewline
\addlinespace[1mm]
Bonn forceps & 0.9726 & \mybar{0.9452}\tabularnewline
\addlinespace[1mm]
hydrodissection canula & 0.9717 & \mybar{0.9434}\tabularnewline
\addlinespace[1mm]
needle holder & 0.9709 & \mybar{0.9418}\tabularnewline
\addlinespace[1mm]
Vannas scissors & 0.9673 & \mybar{0.9346}\tabularnewline
\addlinespace[1mm]
Troutman forceps & 0.9656 & \mybar{0.9312}\tabularnewline
\addlinespace[1mm]
implant injector & 0.9644 & \mybar{0.9288}\tabularnewline
\addlinespace[1mm]
viscoelastic cannula & 0.9545 & \mybar{0.9090}\tabularnewline
\addlinespace[1mm]
Charleux canula & 0.8771 & \mybar{0.7542}\tabularnewline
\addlinespace[1mm]
biomarker & 0.5797 & \mybar{0.1594}\tabularnewline
\bottomrule
\addlinespace[1mm]
\end{tabular}
\end{table}

\section{Conclusion\label{sec:conclusion}}

A surgical tool detection network based on the 50-layer Residual Network
has been trained on the CATARACTS challenge dataset. 
Images extracted from the training videos
are highly correlated and each image is assigned a set of labels instead
of exactly one label. Furthermore, there is an extreme class imbalance.
In consequence, the dataset is heavily preprocessed and augmented.
The class imbalance in combination with the specific distribution
of tool use over the videos made it necessary to validate on a subset
of the data that contains only 16 instead of all 21 classes. Even
worse, it makes it necessary to blindly train the final network without
a validation set.

Different transfer learning settings are compared in a series of experiments.
Fine-tuning ResNet achieves consistently better results than using
ResNet as a fixed feature extractor in combination with a custom classifier.
Next, a specialized classification head to deal with the multi-label
nature of the problem is tested but does not prove useful. However,
using a weighted loss function that increases the loss for underrepresented
classes slightly improves the model's performance.

The resulting network works exceptionally well for some tools but
performance suffers in other cases because not enough training data
is available. Consequently, with a bigger dataset it should be possible
to create a classifier with close to perfect performance. Another possibility 
to improve performance is creating an ensemble of networks starting from
different random initializations or even with different architectures. This might be
especially helpful for the biomarker, as the presumed reason for the bad detection
performance is overfitting.

\bibliographystyle{IEEEtran}
\bibliography{cataracts}

\end{document}